\documentclass[11pt,a4paper]{article}
\usepackage{times,latexsym}
\usepackage{url}
\usepackage[T1]{fontenc}
\usepackage[acceptedWithA]{tacl2018v2}
\usepackage{xspace}
\usepackage{xargs}
\usepackage{xcolor}
\usepackage[colorinlistoftodos,prependcaption,textsize=tiny]{todonotes}
\usepackage{marginnote}
\usepackage{booktabs}
\usepackage{amsmath}
\usepackage{amssymb}
\usepackage{hyperref}
\usepackage{appendix}
\usepackage{multirow}
\usepackage{arydshln}
\usepackage{algorithm}
\usepackage{algpseudocode}

\newcommand\bertlarge{BERT$_{\text{large}}$}
\newcommand\mask{\texttt{[MASK]}}
\newcommand\cls{\texttt{[CLS]}}
\newcommand\sep{\texttt{[SEP]}}
\newcommand\ourmodel{SpanBERT}
\newcommand\sbo{SBO}

\newcommand\googlebert{{Google BERT}}
\newcommand\ourbert{{Our BERT}}
\newcommand\oursinglebert{{Our BERT-1seq}}

\newcommand\ti[1]{\textit{#1}}
\newcommand\tf[1]{\textbf{#1}}
\newcommand\ttt[1]{\texttt{#1}}
\newcommand\mf[1]{\mathbf{#1}}
\usepackage[final]{changes}

\title{SpanBERT: Improving Pre-training by Representing\\ and Predicting Spans}

\author{Mandar Joshi\thanks{~~Equal contribution.} $^{\dagger}$ \quad Danqi Chen$^{* \ddagger\mathsection}$ \quad Yinhan Liu$^{\mathsection}$ \\  { \bf Daniel S. Weld$^{\dagger\epsilon}$ \quad Luke Zettlemoyer$^{\dagger\mathsection}$ \quad Omer Levy$^{\mathsection}$} \\[4pt]
$^{\dagger}$ Allen School of Computer Science \& Engineering, University of Washington, Seattle, WA \\
{\tt \{mandar90,weld,lsz\}@cs.washington.edu}\\[4pt]
$^{\ddagger}$ Computer Science Department, Princeton University, Princeton, NJ \\
{\tt danqic@cs.princeton.edu} \\[4pt]
$^{\epsilon}$Allen Institute of Artificial Intelligence, Seattle\\
{\tt \{danw\}@allenai.org} \\[4pt]
$^{\mathsection}$ Facebook AI Research, Seattle\\
{\tt \{danqi,yinhanliu,lsz,omerlevy\}@fb.com}
}

\begin{document}

\maketitle


\begin{abstract}
We present \ourmodel, a pre-training method that is designed to better represent and predict spans of text.
Our approach extends BERT by (1) masking contiguous random spans, rather than random tokens, and (2) training the span boundary representations to predict the entire content of the masked span, without relying on the individual token representations within it.
\ourmodel\ consistently outperforms BERT and our better-tuned baselines, with substantial gains on span selection tasks such as question answering and coreference resolution.
In particular, with the same training data and model size as \bertlarge, our single model obtains 94.6\% and 88.7\% F1 on SQuAD 1.1 and 2.0 respectively. We also achieve a new state of the art on the OntoNotes coreference resolution task (79.6\% F1), strong performance on the TACRED relation extraction benchmark, 
and even gains on GLUE.\footnote{Our code and pre-trained models are available at \url{https://github.com/facebookresearch/SpanBERT}.}





\end{abstract}


\section{Introduction}

Pre-training methods like BERT~\cite{devlin2018bert} have shown strong performance gains using self-supervised training that masks individual words or subword units.
However, many NLP tasks involve reasoning about relationships between two or more spans of text.
For example, in extractive question answering \cite{rajpurkar2016squad}, determining that the ``Denver Broncos'' is a type of ``NFL team'' is critical for answering the question ``Which NFL team won Super Bowl 50?''
Such spans provide a more challenging target for self supervision tasks, for example predicting ``Denver Broncos'' is much harder than predicting only ``Denver'' when you know the next word is ``Broncos''.
In this paper, we introduce a span-level pretraining approach that consistently outperforms BERT, with the largest gains on span selection tasks such as question answering and coreference resolution.

\begin{figure*}[!t]
\centering
\includegraphics[scale=0.5]{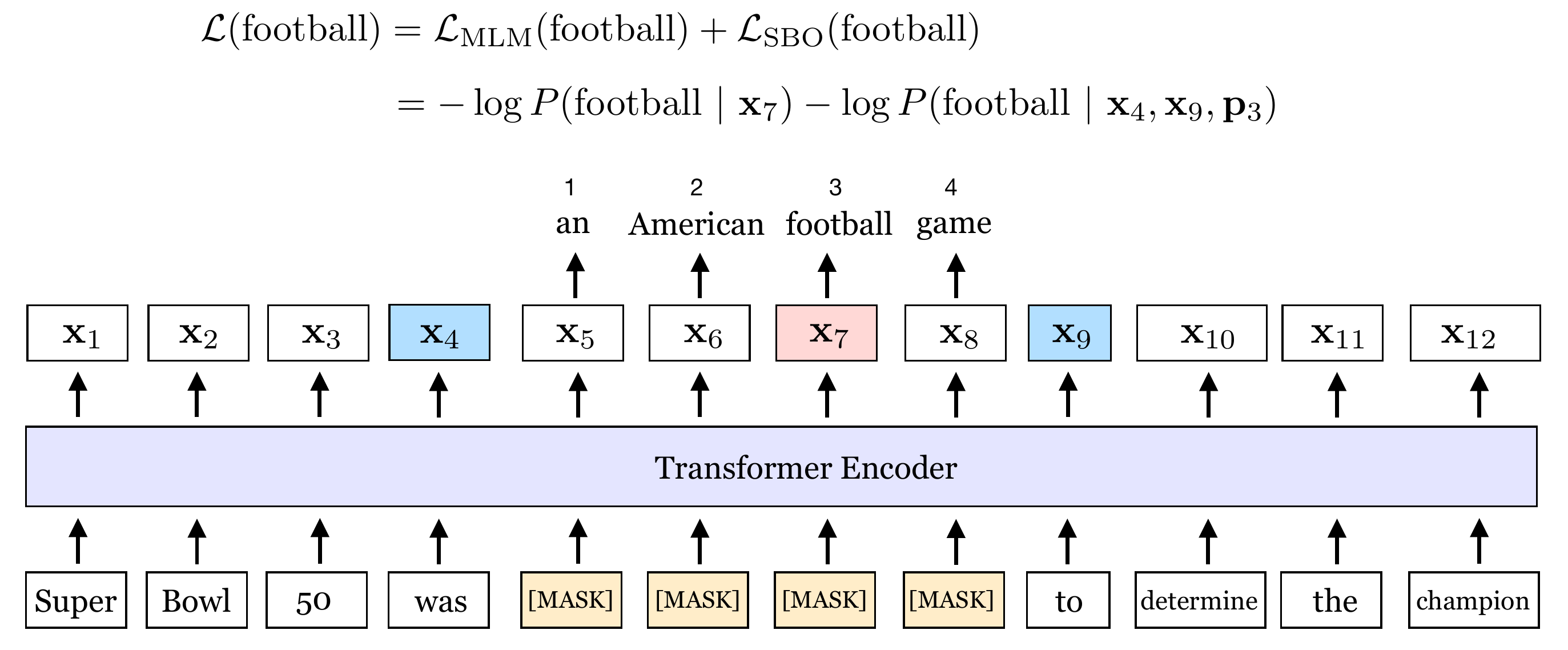}
\caption{An illustration of \ourmodel\ training. The span \ti{an American football game} is masked.
The span boundary objective (SBO) uses the output representations of the boundary tokens, $\mathbf{x}_4$ and $\mathbf{x}_9$ (in blue), to predict each token in the masked span. The equation shows the MLM and SBO loss terms for predicting the  token, \ti{football} (in pink), which as marked by the position embedding $\mathbf{p}_3$, is the \ti{third} token from $x_4$.
}
\label{fig:bertie_spanders}
\end{figure*}

We present \ourmodel, a pre-training method that is designed to better represent and predict spans of text.
Our method differs from BERT in both the masking scheme and the training objectives.
First, we mask random contiguous spans, rather than random individual tokens.
Second, we introduce a novel \emph{span-boundary objective} (SBO) so the model learns to predict the entire masked span from the observed tokens at its boundary.
Span-based masking forces the model to predict entire spans solely using the context in which they appear.
Furthermore, the span-boundary objective encourages the model to store this span-level information at the boundary tokens, which can be easily accessed during the fine-tuning stage. Figure~\ref{fig:bertie_spanders} illustrates our approach.

To implement \ourmodel, we build on a well-tuned replica of BERT, which itself substantially outperforms the original BERT. While building on our baseline,
we find that pre-training on single segments, instead of two half-length segments with the next sentence prediction (NSP) objective, considerably improves performance on most downstream tasks. Therefore, we add our modifications on top of the tuned single-sequence BERT baseline.

Together, our pre-training process yields models that outperform all BERT baselines on a wide variety of tasks, and reach substantially better performance on span selection tasks in particular.
Specifically, our method reaches 94.6\% and 88.7\% F1 on SQuAD 1.1 and 2.0 \cite{rajpurkar2016squad,rajpurkar2018know}, respectively --- reducing error by as much as 27\% compared to our tuned BERT replica.
We also observe similar gains on five additional extractive question answering benchmarks (NewsQA, TriviaQA, SearchQA, HotpotQA, and Natural Questions).\footnote{We use the modified MRQA version of these datasets. See more details in Section~\ref{sec:tasks}.}

\ourmodel\ also arrives at a new state of the art on the challenging CoNLL-2012 (``OntoNotes'') shared task for document-level coreference resolution, where we reach 79.6\% F1, exceeding the previous top model by 6.6\% absolute.
Finally, we demonstrate that \ourmodel\ also helps on tasks that do not explicitly involve span selection, and show that our approach even improves performance on TACRED \cite{zhang2017tacred} and GLUE \cite{wang2019glue}. 

While others show the benefits of adding more data \cite{yang2019xlnet} and increasing model size \cite{lample2019cross}, this work demonstrates the importance of designing good pre-training tasks and objectives, which can also have a remarkable impact.


\section{Background: BERT}
\label{sec:background}

BERT \cite{devlin2018bert} is a self-supervised approach for pre-training a deep transformer encoder \cite{vaswani2017attention}, before fine-tuning it for a particular downstream task.
BERT optimizes two training objectives -- masked language model (MLM) and next sentence prediction (NSP) -- which only require a large collection of unlabeled text.

\paragraph{Notation}
Given a sequence of word or sub-word tokens $X = (x_1, x_2, \ldots, x_n)$, BERT trains an encoder that produces a contextualized vector representation for each token:
$\text{enc}(x_1, x_2, \ldots, x_n) = \mf{x}_1, \mf{x}_2, \ldots, \mf{x}_n$.


\paragraph{Masked Language Model (MLM)}
Also known as a \emph{cloze test}, MLM is the task of predicting missing tokens in a sequence from their placeholders.
Specifically, a subset of tokens $Y \subseteq X$ is sampled and substituted with a different set of tokens.
In BERT's implementation, $Y$ accounts for 15\% of the tokens in $X$; of those, 80\% are replaced with \mask, 10\% are replaced with a random token (according to the unigram distribution), and 10\% are kept unchanged.
The task is to predict the \emph{original} tokens in $Y$ from the modified input.

BERT selects each token in $Y$ independently by randomly selecting a subset.
In \ourmodel, we define $Y$ by randomly selecting \emph{contiguous spans} (Section \ref{sec:span_masking}).

\paragraph{Next Sentence Prediction (NSP)}
The NSP task takes two sequences ($X_A , X_B$) as input, and predicts whether $X_B$ is the direct continuation of $X_A$.
This is implemented in BERT by first reading $X_A$ from the corpus, and then (1) either reading $X_B$ from the point where $X_A$ ended, or (2) randomly sampling $X_B$ from a different point in the corpus. The two sequences are separated by a special \sep token.
Additionally, a special \cls token is added to $X_A , X_B$ to form the input, where the target of \cls is whether $X_B$ indeed follows $X_A$ in the corpus.

\added{In summary, BERT optimizes the MLM and the NSP objectives by masking word pieces uniformly at random in data generated by the bi-sequence sampling procedure. In the next section, we will present our modifications to the data pipeline, masking, and pre-training objectives.}



\section{Model}
\label{sec:model}

We present \ourmodel, a self-supervised pre-training method designed to better represent and predict spans of text.
Our approach is inspired by BERT \cite{devlin2018bert}, but deviates from its bi-text classification framework in three ways.
First, we use a different random process to mask \emph{spans} of tokens, rather than individual ones.
We also introduce a novel auxiliary objective -- the span boundary objective (\sbo) -- which tries to predict the entire masked span using only the representations of the tokens at the span's boundary.
Finally, \ourmodel~samples a single contiguous segment of text for each training example (instead of two), and thus does not use BERT's next sentence prediction objective, which we omit.

\subsection{Span Masking}
\label{sec:span_masking}

\begin{figure}
\centering
\includegraphics[width=\columnwidth]{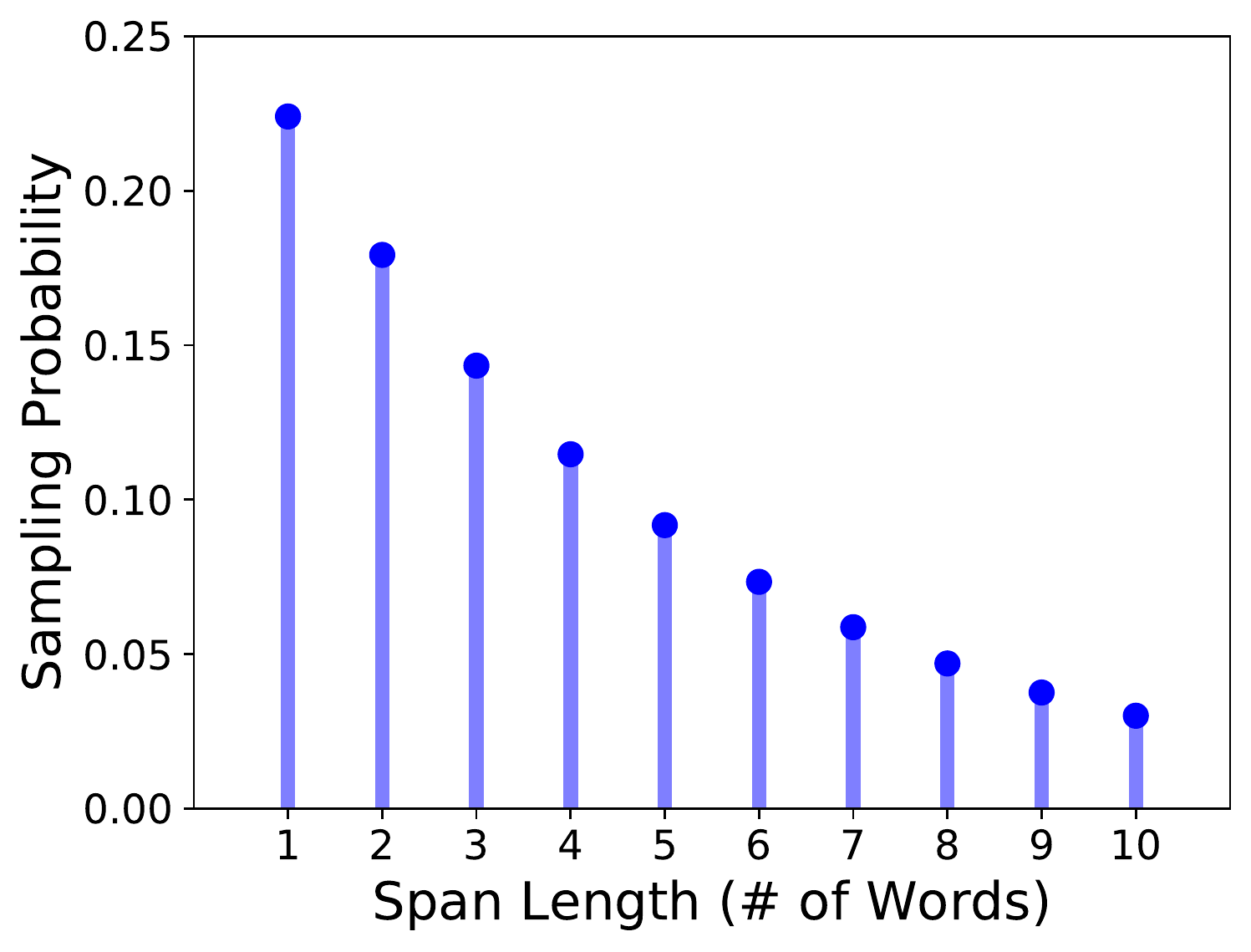}
\caption{We sample random span lengths from a geometric distribution $\ell \sim \mathrm{Geo}(p=0.2)$ clipped at $\ell_{max}=10$.}
\label{fig:geo_dist}
\end{figure}

Given a sequence of tokens $X = ( x_1 , x_2, \ldots , x_n )$, we select a subset of tokens $Y \subseteq X$ by iteratively sampling spans of text until the masking budget (e.g. 15\% of $X$) has been spent.
At each iteration, we first sample a span length (number of words) from a geometric distribution $\ell \sim \mathrm{Geo}(p)$, which is skewed towards shorter spans.
We then randomly (uniformly) select the starting point for the span to be masked. We always sample a sequence of complete words (instead of subword tokens) and the starting point must be the beginning of one word. Following preliminary trials\footnote{\added{We experimented with $p=\{0.1, 0.2, 0.4\}$ and found $0.2$ to perform the best.}}, we set $p=0.2$, and also clip $\ell$ at $\ell_{max}=10$. This yields a mean span length of $\mathrm{mean}(\ell)=3.8$.
Figure~\ref{fig:geo_dist} shows the distribution of span mask lengths.


As in BERT, we also mask 15\% of the tokens in total: replacing 80\% of the masked tokens with \mask, 10\% with random tokens and 10\% with the original tokens. However, we perform this replacement at the span level and not for each token individually; i.e. all the tokens in a span are replaced with \mask or sampled tokens.



\subsection{Span Boundary Objective (SBO)}
\label{sec:span_boundary_objective}

Span selection models \cite{lee2016learning,lee2017end,he2018jointly} typically create a fixed-length representation of a span using its boundary tokens (start and end).
To support such models, we would ideally like the representations for the end of the span to summarize as much of the internal span content as possible.
We do so by introducing a span boundary objective that involves predicting each token of a masked span using only the representations of the observed tokens at the boundaries (Figure~\ref{fig:bertie_spanders}).

Formally, \added{we denote the output of the transformer encoder for each token in the sequence by $\mathbf{x}_1, \ldots, \mathbf{x}_n$}. Given a masked span of tokens $(x_s , \ldots , x_e) \in Y$, where $(s,e)$ indicates its start and end positions, we represent each \added{token $x_i$} in the span using the \added{output encodings of the \ti{external} boundary tokens $\mf{x}_{s-1}$ and $\mf{x}_{e+1}$}, as well as the position embedding of the target token $\mf{p}_{i - s + 1}$:
\begin{align*}
\mf{y}_i = f(\mf{x}_{s-1}, \mf{x}_{e+1}, \mf{p}_{i - s + 1})
\end{align*}
where position embeddings $\mf{p}_1, \mf{p}_2, \ldots$ mark relative positions of the masked tokens with respect to the left boundary token $x_{s-1}$. We implement the representation function $f(\cdot)$ as a 2-layer feed-forward network with GeLU activations \cite{hendrycks2016gelu} and layer normalization~\cite{ba2016layer}:
\begin{eqnarray*}
\mf{h}_0 &=& [\mf{x}_{s-1}; \mf{x}_{e+1}; \mf{p}_{i-s+1}] \\
\mf{h}_1 &=& \mathrm{LayerNorm}(\mathrm{GeLU}(\mf{W}_1 \mf{h}_0 )) \\
\mf{y}_i &=& \mathrm{LayerNorm}(\mathrm{GeLU}(\mf{W}_2 \mf{h}_1 ))
\end{eqnarray*}
We then use the vector representation $\mf{y}_i$ to predict the token $x_i$ and compute the cross-entropy loss exactly like the MLM objective.

\added{\ourmodel~sums the loss from both the span boundary and the regular masked language model objectives for each token $x_i$ in the masked span $(x_s,...,x_e)$, while reusing the input embedding ~\cite{press2017using} for the target tokens in both MLM and SBO:
\begin{eqnarray*}
    \mathcal{L}(x_i) &=& \mathcal{L}_{\mathrm{MLM}}(x_i) + \mathcal{L}_{\mathrm{SBO}}(x_i) \\
    & = & -\log P \left(x_i \mid \mf{x}_i\right) -\log P \left(x_i \mid \mf{y}_i\right)
\end{eqnarray*}
}


\subsection{Single-Sequence Training}
\label{sec:no_nsp}

As described in Section~\ref{sec:background}, BERT's examples contain two sequences of text $(X_A , X_B)$, and an objective that trains the model to predict whether they are connected (NSP).
We find that this setting is almost always worse than simply using a single sequence without the NSP objective (see Section~\ref{sec:results} for further details). We conjecture that single-sequence training is superior to bi-sequence training with NSP because (a) the model benefits from longer full-length contexts, or (b) conditioning on, often unrelated, context from another document adds noise to the masked language model. Therefore, in our approach, we remove both the NSP objective and the two-segment sampling procedure, and simply sample a single contiguous segment of up to $n=512$ tokens, rather than two half-segments that sum up to $n$ tokens together.

\added{In summary, SpanBERT pre-trains span representations by: (1) masking spans of full words using a geometric distribution based masking scheme (Section \ref{sec:span_masking}), (2) optimizing an auxiliary span-boundary objective (Section \ref{sec:span_boundary_objective}) in addition to MLM using a single-sequence data pipeline (Section \ref{sec:no_nsp}). A procedural description can be found in Appendix \ref{pretrain_procedure}.}

\section{Experimental Setup}

\subsection{Tasks}
\label{sec:tasks}

We evaluate on a comprehensive suite of tasks, including seven question answering tasks, coreference resolution, nine tasks in the GLUE benchmark~\cite{wang2019glue}, and relation extraction.
We expect that the span selection tasks, question answering and coreference resolution, will particularly benefit from our span-based pre-training.

\paragraph{Extractive Question Answering}
Given a short passage of text and a question as input, the task of extractive question answering is to select a contiguous span of text in the passage as the answer.

We first evaluate on SQuAD 1.1 and 2.0 ~\cite{rajpurkar2016squad,rajpurkar2018know}, which have served as major question answering benchmarks, particularly for pre-trained models ~\cite{peters2018deep,devlin2018bert,yang2019xlnet}.
We also evaluate on five more datasets from the MRQA shared task~\cite{fisch2019mrqa}\footnote{\href{https://github.com/mrqa/MRQA-Shared-Task-2019}{https://github.com/mrqa/MRQA-Shared-Task-2019}.
MRQA changed the original datasets to unify them into the same format, e.g. all the contexts are truncated to a maximum of 800 tokens and only answerable questions are kept.}: NewsQA~\cite{trischler2017newsqa}, SearchQA~\cite{dunn2017searchqa}, TriviaQA~\cite{joshi2017triviaqa}, HotpotQA~\cite{yang2018hotpotqa} and Natural Questions~\cite{kwiatkowski2019natural}. Because the MRQA shared task does not have a public test set, we split the development set in half to make new development and test sets.
The datasets vary in both domain and collection methodology, making this collection a good testbed for evaluating whether our pre-trained models can generalize well across different data distributions.

Following BERT~\cite{devlin2018bert}, we use the same QA model architecture for all the datasets.
We first convert the passage $P = (p_1, p_2, \ldots, p_{l})$ and question $Q = (q_1, q_2, \ldots, q_{l'})$ 
into a single sequence $X = \cls p_1 p_2 \ldots p_{l} \ttt{[SEP]} q_1 q_2 \ldots q_{l'} \ttt{[SEP]}$, pass it to the pre-trained transformer encoder, and train two linear classifiers independently on top of it for predicting the answer span boundary (start and end). For the unanswerable questions in SQuAD 2.0, we simply set the answer span to be the special token \cls for both training and testing.

\paragraph{Coreference Resolution}
Coreference resolution is the task of clustering mentions in text which refer to the same real-world entities.
We evaluate on the CoNLL-2012 shared task~\cite{pradhan2012conll} for document-level coreference resolution.
\added{We use the \emph{independent} version of the \citet{joshi2019coref} implementation of the higher-order coreference model ~\cite{lee2018higher}. The document is divided into non-overlapping segments of a pre-defined length.\footnote{The length was chosen from \{128, 256, 384, 512\}. See more details in Appendix~\ref{sec:hyperparameters}.} Each segment is encoded independently by the pre-trained transformer encoder, which replaces the original LSTM-based encoder. For each mention span $x$, the model learns a distribution $P(\cdot)$ over possible antecedent spans $Y$}:

\added{
\begin{equation*}
    P(y) = \frac{e^{s(x, y)}}{\sum_{y' \in Y} e^{s(x, y')}}
\end{equation*}
The span pair scoring function $s(x, y)$ is a feedforward neural network over fixed-length span representations and hand-engineered features over $x$ and $y$:
\begin{align*}
    s(x, y) &= s_m(x) + s_m(y) + s_c(x, y) \\
    s_m(x) &= \mathrm{FFNN}_m(\mathbf{g_x}) \\
    s_c(x, y) &= \mathrm{FFNN}_c(\mathbf{g_x}, \mathbf{g_y}, \phi(x, y))
\end{align*}
Here $\mathbf{g_x}$ and $\mathbf{g_y}$ denote the span representations, which are a concatenation of the two transformer output states of the span endpoints and an attention vector computed over the output representations of the token in the span. $\mathrm{FFNN}_m$ and $\mathrm{FFNN}_c$ represent two feedforward neural networks with one hidden layer, and $\phi(x, y)$ represents the hand-engineered features (e.g. speaker and genre information). A more detailed description of the model can be found in~\citet{joshi2019coref}}.


\paragraph{Relation Extraction}
TACRED~\cite{zhang2017tacred} is a challenging relation extraction dataset. Given one sentence and two spans within it -- subject and object -- the task is to predict the relation between the spans from 42 pre-defined relation types, including \ti{no\_relation}.
We follow the entity masking schema from~\citet{zhang2017tacred} and replace the subject and object entities by their NER tags such as  ``\cls \ttt{[SUBJ-PER]} was born in \ttt{[OBJ-LOC]} , Michigan, \ldots'', and finally add a linear classifier on top of the \cls token to predict the relation type.

\paragraph{GLUE}
The General Language Understanding Evaluation (GLUE) benchmark \cite{wang2019glue} consists of 9 sentence-level classification tasks:

\added{\begin{itemize}
    \item Two \emph{sentence-level classification} tasks including CoLA \cite{warstadt2018neural} for evaluating linguistic acceptability and SST-2 \cite{socher2013recursive} for sentiment classification.
    \item Three \emph{sentence-pair similarity} tasks including MRPC \cite{dolan2005automatically}, a binary paraphrasing task sentence pairs from news sources, STS-B~\cite{cer2017semeval}, a graded similarity task for news headlines, and  QQP\footnote{\href{https://data.quora.com/First-Quora-Dataset-Release-Question-Pairs}{https://data.quora.com/First-Quora-Dataset-Release-Question-Pairs}}, a binary paraphrasing tasking between Quora question pairs.
    \item Four \emph{natural language inference} tasks including MNLI \cite{williams2018broad}, QNLI \cite{rajpurkar2016squad}, RTE \cite{dagan2005pascal,bar2006second,giampiccolo2007third} and WNLI \cite{levesque2011winograd}.
\end{itemize}
Unlike question answering, coreference resolution, and relation extraction, these sentence-level tasks do not require \emph{explicit} modeling of span-level semantics. However, they might still benefit from implicit span-based reasoning (e.g., \emph{the Prime Minister} is \emph{the head of the government}).}
\added{Following previous work ~\cite{devlin2018bert,radford2018improving}\footnote{Previous work has excluded WNLI on account of construction issues outlined on the GLUE website -- \url{https://gluebenchmark.com/faq}}, we exclude WNLI from the results to enable a fair comparison}. While recent work \citet{liu2019mtdnn} has applied several task-specific strategies to increase performance on the individual GLUE tasks, we follow BERT's single-task setting and only add a linear classifier on top of the \cls token for these classification tasks.

\subsection{Implementation}
\label{sec:implementation}

We reimplemented BERT's model and pre-training method in \ti{fairseq}~\cite{ott2019fairseq}.
We used the model configuration of \bertlarge\ as in \newcite{devlin2018bert} and also pre-trained all our models on the same corpus: BooksCorpus and English Wikipedia using \textit{cased} Wordpiece tokens.

Compared to the original BERT implementation, the main differences in our implementation include: (a) We use different masks at each epoch while BERT samples 10 different masks for each sequence during data processing. (b) We remove all the short-sequence strategies used before (they sampled shorter sequences with a small probability 0.1; they also first pre-trained with smaller sequence length of 128 for 90\% of the steps). Instead, we always take sequences of up to 512 tokens until it reaches a document boundary. \added{We refer readers to \newcite{liu2019roberta} for further discussion on these modifications and their effects.}


As in BERT, the learning rate is warmed up over the first 10,000 steps to a peak value of
1e-4, and then linearly decayed. \added{We retain $\beta$ hyperparameters ($\beta_1 = 0.9$, $\beta_2= 0.999$) and a decoupled weight decay ~\cite{loshchilov2018decoupled} of $0.1$. We also keep a dropout of 0.1 on all layers and attention weights, and a GeLU activation function~\cite{hendrycks2016gelu}}. We deviate from the optimization by running for 2.4M steps and using an epsilon of 1e-8 for AdamW \cite{kingma2014adam}, which converges to a better set of model parameters.  \added{Our implementation uses a batch size of 256 sequences with a maximum of 512 tokens.\footnote{On the average, this is approximately 390 sequences since some documents have fewer than 512 tokens}} For the SBO, we use 200 dimension position embeddings $\mf{p}_1, \mf{p}_2, \ldots$ to mark positions relative to the left  boundary  token. The pre-training was done on 32 Volta V100 GPUs and took 15 days to complete.

Fine-tuning is implemented based on HuggingFace's codebase ~\cite{Wolf2019HuggingFacesTS}
and more details are given in Appendix~\ref{sec:hyperparameters}.

\subsection{Baselines}
\label{sec:baselines}

We compare \ourmodel~to three baselines:

\paragraph{Google BERT} The pre-trained models released by \newcite{devlin2018bert}.\footnote{\href{https://github.com/google-research/bert}{https://github.com/google-research/bert}.}

\paragraph{Our BERT}
Our reimplementation of BERT with improved data preprocessing and optimization (Section~{\ref{sec:implementation}}).

\paragraph{Our BERT-1seq}
Our reimplementation of BERT trained on single full-length sequences without NSP (Section~{\ref{sec:no_nsp}}).

\begin{table}[!t]
  \centering
  \small
  \begin{tabular}{l c c c c c c}
    \toprule
     & & \multicolumn{2}{c}{SQuAD 1.1} & & \multicolumn{2}{c}{SQuAD 2.0} \\
     & & EM & F1 & & EM & F1 \\
    \midrule
    Human Perf. & & 82.3 & 91.2 & & 86.8 & 89.4 \\
    \midrule
    \googlebert & & 84.3 & 91.3 & & 80.0 &	83.3 \\
    \ourbert & & 86.5 & 92.6 & & 82.8 &	85.9 \\
    \oursinglebert & & 87.5 & 93.3 & & 83.8 & 86.6 \\
    \ourmodel & & \tf{88.8} & \tf{94.6} & & \tf{85.7} & \tf{88.7} \\
    \bottomrule
  \end{tabular}
  \caption{Test results on SQuAD 1.1 and SQuAD 2.0.}
  \label{tab:squad-results}
\end{table}





\section{Results}
\label{sec:results}

\begin{table*}[!t]
  \centering
  \small
  \begin{tabular}{l @{\hspace{0.5cm}} c c c c c @{\hspace{0.5cm}} c}
    \toprule
     & {NewsQA} & {TriviaQA} & {SearchQA}	& {HotpotQA} & {Natural Questions} & Avg. \\
    \midrule
    \googlebert & 68.8 & 77.5 & 81.7 & 78.3 & 79.9 &  77.3 \\
    \ourbert & 71.0 & 79.0 & 81.8 & 80.5 & 80.5 & 78.6 \\
    \oursinglebert & 71.9 & 80.4 & 84.0 & 80.3 &  81.8 & 79.7 \\
    \ourmodel & \tf{73.6} & \tf{83.6} & \tf{84.8} &  \tf{83.0} & \tf{82.5} & \tf{81.5} \\
    \bottomrule
  \end{tabular}
  \caption{Performance (F1) on the five MRQA extractive question answering tasks.}
  \label{tab:mrqa-results}
\end{table*}

\begin{table*}[!t]
\small
\centering
\setlength{\tabcolsep}{4pt}
\begin{tabular}{l@{\hspace{0.5cm}}ccc@{\hspace{0.5cm}}ccc@{\hspace{0.5cm}}ccc@{\hspace{0.5cm}}c}
\toprule
 & \multicolumn{3}{c}{MUC}& \multicolumn{3}{c}{$\text{B}^3$}&\multicolumn{3}{c}{$\text{CEAF}_{\phi_4}$} \\
& P & R & F1 & P & R & F1 & P & R & F1 & Avg. F1 \\
\midrule
Prev. SotA: \cite{lee2018higher}  & 81.4 & 79.5 & 80.4 & 72.2 & 69.5 & 70.8 & 68.2 & 67.1 & 67.6 & 73.0 \\
\midrule
\googlebert & 84.9 & 82.5 & 83.7 & 76.7 & 74.2 & 75.4 & 74.6 & 70.1 & 72.3 & 77.1 \\
\ourbert & 85.1 & 83.5 & 84.3 & 77.3 & 75.5 & 76.4 & 75.0 & 71.9 & 73.9 & 78.3 \\
\oursinglebert & 85.5 & 84.1 & 84.8 & 77.8 & 76.7 & 77.2 & 75.3 & 73.5 & 74.4 & 78.8 \\
\ourmodel &  \tf{85.8} & \tf{84.8} & \tf{85.3} & \tf{78.3} & \tf{77.9} & \tf{78.1} & \tf{76.4} & \tf{74.2} & \tf{75.3} & \tf{79.6}  \\
\bottomrule
\end{tabular}
\caption{Performance on the OntoNotes coreference resolution benchmark. The main evaluation is the average F1 of three metrics: MUC, $\text{B}^3$, and $\text{CEAF}_{\phi_4}$ on the test set.}
\label{tab:coref-results}
\end{table*}

\begin{table}[!t]
\small
\centering
\setlength{\tabcolsep}{4pt}
\begin{tabular}{l@{\hspace{0.5cm}}ccc}
\toprule
& P & R & F1 \\
\midrule
BERT$_\text{EM}$~\cite{Soares2019matching} & - & - & 70.1 \\
BERT$_\text{EM}$+MTB$^*$ & - & - & \tf{71.5} \\
\midrule
\googlebert & 69.1 & 63.9 & 66.4 \\
\ourbert & 67.8 & 67.2 & 67.5 \\
\oursinglebert & \tf{72.4} & 67.9 & 70.1 \\
\ourmodel & 70.8 & \tf{70.9} & \tf{70.8} \\
\bottomrule
\end{tabular}
\caption{Test performance on the TACRED relation extraction benchmark. BERT$_\text{EM}$ and BERT$_\text{EM}$+MTB from \citet{Soares2019matching} are the current state-of-the-art. $^*$: BERT$_\text{EM}$+MTB incorporated an intermediate ``matching the blanks'' pre-training on the entity-linked text based on English Wikipedia, which is not a direct comparison to ours trained only from raw text. }
\label{tab:tacred-results}
\end{table}

\begin{table*}[!t]
  \centering
  \small
  \begin{tabular}{l @{\hspace{0.5cm}} c c c c c c c c @{\hspace{0.5cm}} c}
    \toprule
     & {CoLA} & {SST-2} & {MRPC} & {STS-B} & {QQP} & MNLI & {QNLI} & {RTE} & (Avg) \\
    \midrule
    \googlebert & 59.3 & \textbf{95.2} & 88.5/84.3 & 86.4/88.0 & 71.2/89.0 & 86.1/85.7 & 93.0 & 71.1 & 80.4\\
    \ourbert & 58.6 & 93.9 & 90.1/86.6 & 88.4/89.1 & 71.8/89.3 & 87.2/86.6 & 93.0 & 74.7 & 81.1 \\
    \oursinglebert & 63.5 & 94.8 & \tf{91.2}/87.8 & 89.0/88.4 & \tf{72.1}/\tf{89.5} & 88.0/87.4 & 93.0 & 72.1 & 81.7 \\
    \ourmodel & \tf{64.3} & 94.8 & 90.9/\tf{87.9} & \tf{89.9}/\tf{89.1} & 71.9/\tf{89.5} & \tf{88.1}/\tf{87.7} & \tf{94.3} & \tf{79.0} & \textbf{82.8}\\
    \bottomrule
  \end{tabular}
  \caption{Test set performance on GLUE tasks. MRPC: F1/accuracy, STS-B: Pearson/Spearmanr correlation, QQP: F1/accuracy, MNLI: matched/mistached accuracies and accuracy for all the other tasks. WNLI (not shown) is always set to majority class (65.1\% accuracy) and included in the average.}
  \label{tab:glue-results}
\end{table*}

We compare \ourmodel\ to the baselines per task, and draw conclusions based on the overall trends.

\subsection{Per-Task Results}

\paragraph{Extractive Question Answering}
Table~\ref{tab:squad-results} shows the performance on both SQuAD 1.1 and 2.0.
\ourmodel\ exceeds our BERT baseline by 2.0\% and 2.8\% F1 respectively (3.3\% and 5.4\% over \googlebert). In SQuAD 1.1, this result accounts for over 27\% error reduction, reaching 3.4\% F1 \emph{above} human performance.

Table~\ref{tab:mrqa-results} demonstrates that this trend goes beyond SQuAD, and is consistent in every MRQA dataset.
On average, we see a 2.9\% F1 improvement from our reimplementation of BERT.
Although some gains are coming from single-sequence training (+1.1\%), most of the improvement stems from span masking and the span boundary objective (+1.8\%), with particularly large gains on TriviaQA (+3.2\%) and HotpotQA (+2.7\%).

\paragraph{Coreference Resolution}
Table~\ref{tab:coref-results} shows the performance on the OntoNotes coreference resolution benchmark. Our BERT reimplementation improves the Google BERT model by 1.2\% on the average F1 metric and single-sequence training brings another 0.5\% gain. Finally, \ourmodel\  improves considerably on top of that, achieving a new state of the art of 79.6\% F1 (previous best result is 73.0\%).

\paragraph{Relation Extraction}
Table~\ref{tab:tacred-results} shows the performance on TACRED.
\ourmodel\ exceeds our reimplementation of BERT by 3.3\% F1 and achieves close to the current state of the art ~\cite{Soares2019matching} --- Our model performs better than their BERT$_\text{EM}$ but is 0.7 point behind BERT$_\text{EM}$ + MTB which used entity-linked text for additional pre-training. Most of this gain (+2.6\%) stems from single-sequence training although the contribution of span masking and the span boundary objective is still a considerable 0.7\%, resulting largely from higher recall.


\paragraph{GLUE}
Table~\ref{tab:glue-results} shows the performance on GLUE.
For most tasks, the different models appear to perform similarly.
Moving to single-sequence training without the NSP objective substantially improves CoLA, and yields smaller (but considerable) improvements on MRPC and MNLI.
The main gains from \ourmodel\ are in the SQuAD-based QNLI dataset (+1.3\%) and in RTE (+6.9\%), the latter accounting for most of the rise in \ourmodel's GLUE average.

\subsection{Overall Trends}

We compared our approach to three BERT baselines on 17 benchmarks, and found that \textbf{\ourmodel\ outperforms BERT on almost every task}.
In 14 tasks, \ourmodel\ performed better than all baselines.
In 2 tasks (MRPC and QQP), it performed on-par in terms of accuracy with single-sequence trained BERT, but still outperformed the other baselines.
In one task (SST-2), Google's BERT baseline performed better than \ourmodel\ by 0.4\% accuracy.

When considering the magnitude of the gains, it appears that \textbf{\ourmodel\ is especially better at extractive question answering}.
In SQuAD 1.1, for example, we observe a solid gain of 2.0\% F1 even though the baseline is already well above human performance.
On MRQA, \ourmodel\ improves between 2.0\% (Natural Questions) and 4.6\% (TriviaQA) F1 on top of our BERT baseline.

Finally, we observe that \textbf{single-sequence training works considerably better than bi-sequence training with next sentence prediction (NSP)} with BERT's choice of sequence lengths for a wide variety of tasks .
This is surprising because BERT's ablations showed gains from the NSP objective \cite{devlin2018bert}.
However, the ablation studies still involved bi-sequence data processing, i.e. the pre-training stage only controlled for the NSP objective while still sampling two half-length sequences.
We hypothesize that bi-sequence training, as it is implemented in BERT (see Section~\ref{sec:background}), impedes the model from learning longer-range features, and consequently hurts performance on many downstream tasks.


\section{Ablation Studies}

\begin{table*}[!t]
    \centering
    \small
    \begin{tabular}{l @{\hspace{0.2cm}} c c c c c c c}
    \toprule
    &  SQuAD 2.0 &  NewsQA & TriviaQA & Coreference & MNLI-m & QNLI & GLUE (Avg)\\
    \midrule
    Subword Tokens & 83.8 &	72.0 &	76.3 & \tf{77.7} &	86.7 & 92.5 & 83.2\\
    Whole Words  &	84.3 & 72.8 & 77.1 & 76.6 & 86.3 & 92.8  & 82.9\\
    Named Entities &	84.8 &	72.7 &	78.7 & 75.6 &	86.0 & 93.1  & 83.2\\
    Noun Phrases &	85.0 &	\tf{73.0} &	77.7 & 76.7 &	86.5	& 93.2  & \tf{83.5}\\
    Geometric Spans & \tf{85.4} &	\tf{73.0} &	\tf{78.8} & 76.4 & \tf{87.0} & \tf{93.3}  & 83.4\\
    \bottomrule
    \end{tabular}
    \caption{The effect of replacing BERT's original masking scheme (Subword Tokens) with different masking schemes. Results are F1 scores for QA tasks and accuracy for MNLI and QNLI on the development sets. All the models are based on bi-sequence training with NSP.}
    \label{tab:masking-schemes}
\end{table*}

\begin{table*}[!t]
    \centering
    \small
    \begin{tabular}{l @{\hspace{0.2cm}} c c c c c c c}
    \toprule
    &  SQuAD 2.0 & NewsQA & TriviaQA & Coref &  MNLI-m & QNLI & GLUE (Avg) \\
    \midrule
    Span Masking (2seq) + NSP & 85.4 &	73.0 &	78.8 & 76.4  &	87.0 & 93.3 & 83.4\\
    Span Masking (1seq) & 86.7 & 73.4 &	80.0 & 76.3  & 87.3 & 93.8 & 83.8\\
    Span Masking (1seq) + SBO & \tf{86.8} &	\tf{74.1} &	\tf{80.3} &	\tf{79.0} & \tf{87.6} & \tf{93.9} & \tf{84.0}\\
    \bottomrule
    \end{tabular}
    \caption{The effects of different auxiliary objectives, given MLM over random spans as the primary objective.}
    \label{tab:sbo-ablation}
\end{table*}

We compare our random span masking scheme with linguistically-informed masking schemes, and find that masking random spans is a competitive and often better approach.
We then study the impact of the span boundary objective (SBO), and contrast it with BERT's next sentence prediction (NSP) objective.\footnote{To save time and resources, we use the checkpoints at 1.2M steps for all the ablation experiments.}

\subsection{Masking Schemes}

Previous work ~\cite{sun2019ernie} has shown improvements in downstream task performance by masking linguistically-informed spans during pre-training for Chinese data.
We compare our random span masking scheme with masking of linguistically-informed spans. 
Specifically, we train the following five baseline models differing only in the way tokens are masked.

\paragraph{Subword Tokens} We sample random Wordpiece tokens, as in the original BERT.

\paragraph{Whole Words}
We sample random words, and then mask all of the subword tokens in those words. The total number of masked subtokens is around 15\%.

\paragraph{Named Entities}
At 50\% of the time, we sample from named entities in the text, and sample random whole words for the other 50\%. The total number of masked subtokens is 15\%. Specifically, we run spaCy's named entity recognizer\footnote{\url{https://spacy.io/}} on the corpus and select all the non-numerical named entity mentions as candidates.

\paragraph{Noun Phrases}
Similar as \ti{Named Entities}, we sample from noun phrases at 50\% of the time. The noun phrases are extracted by running spaCy's constituency parser.

\paragraph{Geometric Spans}
We sample random spans from a geometric distribution, as in our \ourmodel\ (see Section \ref{sec:span_masking}).

\vspace{5pt}
 Table~\ref{tab:masking-schemes} shows how different pre-training masking schemes affect performance on the development set of a selection of tasks. \added{All the models are evaluated on the development sets and are based on the default BERT setup of bi-sequence training with NSP; the results are not directly comparable to the main evaluation.}
With the exception of coreference resolution, masking random spans is preferable to other strategies.
Although linguistic masking schemes (named entities and noun phrases) are often competitive with random spans, their performance is not consistent; for instance, masking noun phrases achieves parity with random spans on NewsQA, but underperforms on TriviaQA (-1.1\% F1).

On coreference resolution, we see that masking random subword tokens is preferable to any form of span masking.
Nevertheless, we shall see in the following experiment that combining random span masking with the span boundary objective can improve upon this result considerably.

\subsection{Auxiliary Objectives}

In Section~\ref{sec:results}, we saw that bi-sequence training with the next sentence prediction (NSP) objective can hurt performance on downstream tasks, when compared to single-sequence training.
We test whether this holds true for models pre-trained with span masking, and also evaluate the effect of replacing the NSP objective with the span boundary objective (SBO).

Table~\ref{tab:sbo-ablation} confirms that single-sequence training typically improves performance. Adding SBO further improves performance, with a substantial gain on coreference resolution (+2.7\% F1) over span masking alone.
Unlike the NSP objective, SBO does not appear to have any adverse effects.



\section{Related Work}

Pre-trained contextualized word representations that can be trained from unlabeled text ~\cite{dai2015semi,melamud2016context2vec,peters2018deep} have had immense impact on NLP lately,
particularly as methods for initializing a large model before fine-tuning it for a specific task ~\cite{howard2018universal,radford2018improving,devlin2018bert}.
Beyond differences in model hyperparameters and corpora, these methods mainly differ in their pre-training tasks and loss functions, with a considerable amount of contemporary literature proposing augmentations of BERT's masked language modeling (MLM) objective.

While previous and concurrent work has looked at masking ~\cite{sun2019ernie} or dropping ~\cite{song2019mass,chan2019kermit} multiple words from the input -- particularly as pretraining for language generation tasks -- \ourmodel\  pretrains span representations ~\cite{lee2016learning}, which are widely used for question answering, coreference resolution and a variety of other tasks.
ERNIE ~\cite{sun2019ernie} shows improvements on Chinese NLP tasks using phrase and named entity masking.
MASS ~\cite{song2019mass} focuses on language generation tasks, and adopts the encoder-decoder framework to reconstruct a sentence fragment given the remaining part of the sentence.
We attempt to more explicitly model spans using the SBO objective, and show that (geometrically distributed) random span masking works as well, and sometimes  better than, masking linguistically-coherent spans.
We evaluate on English benchmarks for question answering, relation extraction, and coreference resolution in addition to GLUE.

A different ERNIE ~\cite{zhang2019ernie} focuses on integrating structured knowledge bases with contextualized representations with an eye on knowledge-driven tasks like entity typing and relation classification.
UNILM ~\cite{dong2019unified} uses multiple language modeling objectives -- unidirectional (both left-to-right and right-to-left), bidirectional, and sequence-to-sequence prediction -- to aid generation tasks like summarization and question generation.
XLM \cite{lample2019cross} explores cross-lingual pre-training for multilingual tasks such as translation and cross-lingual classification.
Kermit \cite{chan2019kermit}, an insertion based  approach, fills in missing tokens (instead of predicting masked ones) during pretraining; they show improvements on machine translation and zero-shot question answering.

Concurrent with our work, RoBERTa ~\cite{liu2019roberta} presents a replication study of BERT pre-training that measures the impact of many key hyperparameters and training data size. Also concurrent, XLNet ~\cite{yang2019xlnet} combines an autoregressive loss and the Transformer-XL ~\cite{dai2019transformer} architecture with a more than an eight-fold increase in data to achieve current state-of-the-art results on multiple benchmarks.
XLNet also masks spans (of 1-5 tokens) during pre-training, but predicts them autoregressively.
Our model focuses on incorporating span-based pre-training, and as a side effect, we present a stronger BERT baseline while controlling for the corpus, architecture, and the number of parameters.

Related to our SBO objective, \ti{pair2vec} ~\cite{joshi2019pair2vec} encodes word-pair relations using a negative sampling-based multivariate objective during pre-training.
Later, the word-pair representations are injected into the attention-layer of downstream tasks, and thus encode limited downstream context.
Unlike pair2vec, our SBO objective yields ``pair'' (start and end tokens of spans) representations which more fully encode the context during both pre-training and finetuning, and are thus more appropriately viewed as \emph{span} representations. \citet{Stern2018BlockwisePD} focus on improving language generation speed using a block-wise parallel decoding scheme; they make predictions for multiple time steps in parallel and then back off to the longest prefix validated by a scoring model. Also related are sentence representation methods ~\cite{Kiros2015SkipThoughtV,Logeswaran2018AnEF} which focus on predicting surrounding contexts from sentence embeddings.



\section{Conclusion}

We presented a new method for span-based pre-training which extends BERT by (1) masking contiguous random spans, rather than random tokens, and (2) training the span boundary representations to predict the entire content of the masked span, without relying on the individual token representations within it. 
Together, our pre-training process yields models that outperform all BERT baselines on a variety of tasks, and reach substantially better performance on span selection tasks in particular.

\section*{Acknowledgements}

We would like to thank Pranav Rajpurkar and Robin Jia for patiently helping us evaluate \ourmodel\ on SQuAD.
We thank the anonymous reviewers, the action editor, and our colleagues at Facebook AI Research and the University of Washington for their insightful feedback that helped improve the paper.

\bibliography{references}
\bibliographystyle{acl_natbib}

\begin{appendices}

\appendix
\section*{Appendices}
\addcontentsline{toc}{section}{Appendices}
\renewcommand{\thesubsection}{\Alph{subsection}}

\added{
\subsection{Pre-training Procedure}
\label{pretrain_procedure}
We describe our pre-training procedure as follows:
\begin{enumerate}
    \item Divide the corpus into single contiguous blocks of up to 512 tokens.
    \item At each step of pre-training:
    \begin{enumerate}
        \item Sample a batch of blocks uniformly at random.
        \item Mask 15\% of word pieces in each block in the batch using the span masking scheme (Section \ref{sec:span_masking}).
        \item For each masked token $x_i$, optimize $\mathcal{L}(x_i) = \mathcal{L}_{\text{MLM}}(x_i) + \mathcal{L}_{\text{SBO}}(x_i)$ (Section \ref{sec:span_boundary_objective}).
    \end{enumerate}
\end{enumerate}
}

\subsection{Fine-tuning Hyperparameters}
\label{sec:hyperparameters}

We apply the following fine-tuning hyperparameters to all methods, including the baselines.

\paragraph{Extractive Question Answering}
For all the question answering tasks, we use \texttt{max\_seq\_length} = 512 and a sliding window of size $128$ if the lengths are longer than 512. We choose learning rates from \{5e-6, 1e-5, 2e-5, 3e-5, 5e-5\} and batch sizes from \{16, 32\} and fine-tune 4 epochs for all the datasets.

\paragraph{Coreference Resolution}
We divide the documents into multiple chunks of lengths up to \texttt{max\_seq\_length} and encode each chunk independently. We choose \texttt{max\_seq\_length} from \{128, 256, 384, 512\}, BERT learning rates from \{1e-5, 2e-5\}, task-specific learning rates from \{1e-4, 2e-4, 3e-4\} and fine-tune 20 epochs for all the datasets. We use batch size $=1$ (one document) for all the experiments.

\paragraph{TACRED/GLUE} We use \texttt{max\_seq\_length} = 128 and choose learning rates from \{5e-6, 1e-5, 2e-5, 3e-5, 5e-5\} and batch sizes from \{16, 32\} and fine-tuning 10 epochs for all the datasets. The only exception is CoLA, where we used 4 epochs (following \newcite{devlin2018bert}), because 10 epochs lead to severe overfitting.

\end{appendices}

\end{document}